\documentclass[conference]{IEEEtran}
\IEEEoverridecommandlockouts

\usepackage{cite}
\usepackage{amsmath,amssymb,amsfonts}
\usepackage{algorithmic}
\usepackage{graphicx}
\usepackage{textcomp}
\usepackage{xcolor}
\usepackage{booktabs}   
\usepackage{multirow}   
\usepackage{makecell}   
\usepackage{array}      
\usepackage{hyperref}   

\usepackage[ruled,vlined]{algorithm2e}
\usepackage[T1]{fontenc}

\def\BibTeX{{\rm B\kern-.05em{\sc i\kern-.025em b}\kern-.08em
    T\kern-.1667em\lower.7ex\hbox{E}\kern-.125emX}}

\begin{document}

\title{Clore: Interactive Pathology Image Segmentation with Click-based Local Refinement\thanks{\IEEEauthorrefmark{3}Corresponding authors.}}
\author{
    \IEEEauthorblockN{
        Tiantong Wang\IEEEauthorrefmark{1}, 
        Minfan Zhao\IEEEauthorrefmark{1}, 
        Jun Shi\IEEEauthorrefmark{1}\IEEEauthorrefmark{3}, 
        Hannan Wang\IEEEauthorrefmark{1} and 
        Yue Dai\IEEEauthorrefmark{2}
    }
    \IEEEauthorblockA{
        \IEEEauthorrefmark{1}School of Computer Science and Technology, University of Science and Technology of China\\
        \IEEEauthorrefmark{2}School of Artificial Intelligence and Data Science, University of Science and Technology of China
    }
}

\maketitle
\begin{abstract}
Recent advancements in deep learning-based interactive segmentation methods have significantly improved pathology image segmentation. Most existing approaches utilize user-provided positive and negative clicks to guide the segmentation process. However, these methods primarily rely on iterative global updates for refinement, which lead to redundant re-prediction and often fail to capture fine-grained structures or correct subtle errors during localized adjustments. To address this limitation, we propose the \textbf{C}lick-based \textbf{Lo}cal \textbf{Re}finement (\textbf{Clore}) pipeline, a simple yet efficient method designed to enhance interactive segmentation. The key innovation of Clore lies in its hierarchical interaction paradigm: the initial clicks drive global segmentation to rapidly outline large target regions, while subsequent clicks progressively refine local details to achieve precise boundaries. This approach not only improves the ability to handle fine-grained segmentation tasks but also achieves high-quality results with fewer interactions. Experimental results on four datasets demonstrate that Clore achieves the best balance between segmentation accuracy and interaction cost, making it an effective solution for efficient and accurate interactive pathology image segmentation. Code will be released at \href{https://github.com/legend5661/Clore.git}{https://github.com/legend5661/Clore.git}.
\end{abstract}


\section{Introduction}
\label{sec:intro}
Pathology image-based examination has become the gold standard for tumor diagnosis~\cite{litjens2017survey}, where pathology image segmentation plays a pivotal role in the examination procedure. Accurate segmentation of Regions of Interest (ROIs), such as tumor boundaries, is essential for cancer staging and treatment planning. While deep learning has advanced automated segmentation, traditional fully supervised methods require labor-intensive pixel-level annotations, posing challenges due to high labeling costs for complex pathology images ~\cite{wang2022medical}. Weakly supervised approaches like using scribbles and bounding boxes~\cite{zhang2022weakly,lu2021data,fan2024pathmamba} reduce annotation burdens but suffer from accuracy degradation (8\%–12\% Dice score decline in tumor subtype tasks~\cite{bearman2016s}). Semi-supervised methods face limitations from noisy pseudo-labels~\cite{zhong2024semi,sun2024semi}, struggling to meet higher precision demands.

Interactive segmentation leverages the complementary strengths of human expertise and deep learning to achieve accurate segmentation with minimal manual effort. Existing methods allow users to provide guidance through various forms of interaction, such as clicks, bounding boxes, scribbles, and masks~\cite{liew2021deep,xu2017deep,ling2019fast,liu2024rethinking}. While early approaches focused on designing interaction modalities, recent research has shifted toward improving the efficiency of utilizing these interactive signals. A number of studies have sought to enhance interactive segmentation performance from different perspectives, including better propagation and refinement of user inputs to ensure precise segmentation results~\cite{chen2021conditional,min2023cgam,li2024interactive,lee2024sn}.

 Due to the high precision requirements for pathology image segmentation, the amount of interactive information needed far exceeds that required in natural image segmentation. Most existing methods reapply all interaction signals to the entire image per iteration, leading to redundant re-prediction and high computational costs. Meanwhile, their reliance on iterative global updates often overlooks fine-grained structures and localized corrections. Methods like FocalClick~\cite{chen2022focalclick} introduce a two-step strategy (coarse segmentation + local refinement) by employing an auxiliary refinement network to solve this problem, but they also raise new issues: applying the two-step strategy in every interaction disregards user intent to some extent and incurs additional computational burden. Early interactions are typically designed to cover broad targets with minimal clicks, often serving as global spatial priors, whereas premature local refinement may overemphasize localized features, conflict with foreground–background classification, and degrade performance. In contrast, later-stage interactions focus on precise local corrections, where refinement better aligns with user intent. However, prior methods have largely overlooked the distinct prior information inherent in different interaction phases. Moreover, using extra auxiliary refinement networks will inflates computational overhead and introduce additional difficulties during model training. \textit{These limitations in both interaction-stage handling and refinement network design pose significant challenges to realizing accurate and efficient pathology image segmentation.}
 \begin{figure*}[ht]
    \centering
    \includegraphics[width=0.94\textwidth]{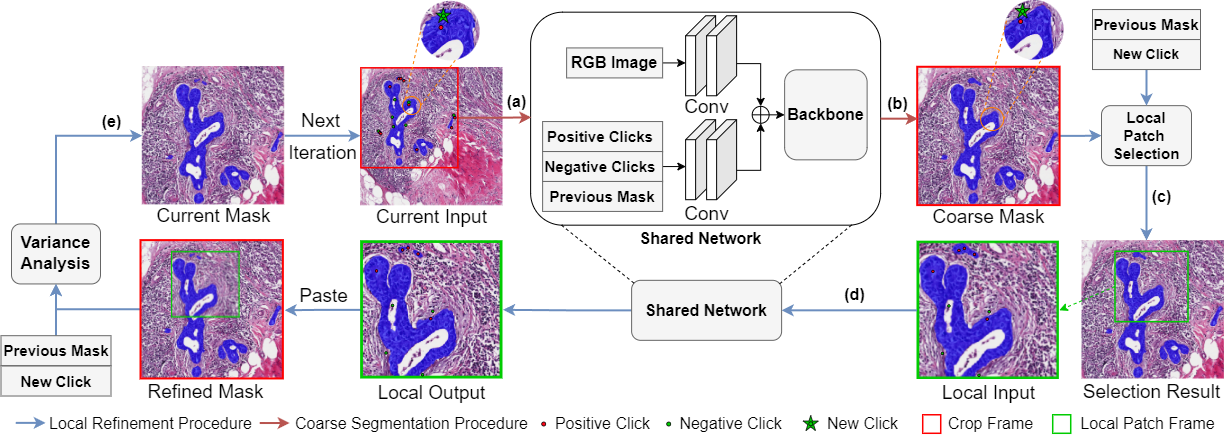}
    \caption{An overview of the proposed method}
    \label{figoverview}
\end{figure*}

To address these challenges, we propose Interactive pathology Image Segmentation with \textbf{C}lick-based \textbf{Lo}cal \textbf{Re}finement (\textbf{Clore}). The method employs a customizable triggered local refinement strategy under click-based interaction to enable efficient pathology image segmentation. We also employ a shared segmentation network for both global coarse segmentation and local refinement. Specifically, users can set a hyperparameter $n$, where $n$ is a positive integer, it indicates the click number after which the Local Refinement is triggered. For the first $n$ clicks,  our method directly generates segmentation results through efficient global prediction. After $n$ click iterations, the coarse global segmentation will be followed by the selection of a Local Patch (\textbf{LP}) via a selection algorithm. The LP typically corresponds to the most critical region of interest, which is then refined using the same segmentation network. Finally, a merge algorithm integrates the refined mask with the previous segmentation mask to produce the final output. Clore method achieves more efficient fusion of global and local information, this enables the approach to achieve superior segmentation details, particularly in pathology images, which typically exhibit high resolution.

\section{Methodology}
\label{sec:method}
\subsection{Overview}
Figure~\ref{figoverview} outlines our Click-based Local Refinement method for pathology image segmentation. The segmentation pipeline operates in two distinct phases depending on the number of user clicks.
For the initial $n$ clicks, the process follows a coarse prediction loop. This iterative loop consists of steps (a) and (b). In step (a), we crop the original image to a smaller region with a higher foreground ratio based on the prediction from the previous iteration and the current click. Notably, this cropping step is skipped for the very first click. Subsequently, in step (b), the cropped image is resized to the network's input size of $512 \times 512$. It is then passed to the segmentation network, along with the encoded clicks and the previous mask, to generate a coarse global prediction mask, $M_{\textnormal{c}}$. In this initial phase, this coarse mask serves as the final output for the current iteration and is used as the basis for the cropping in the next iteration's step (a).
After more than $n$ clicks have been provided, the method transitions into a refinement phase by incorporating steps (c), (d), and (e) into the loop. The process begins as before with steps (a) and (b) to obtain a coarse prediction mask $M_{\textnormal{c}}$. Following this, in step (c), we generate a Local Patch (LP) based on the current coarse prediction $M_{\textnormal{c}}$, the prediction mask from the previous iteration $M_{\textnormal{p}}$, and the newly added click $C_{\textnormal{n}}$. In step (d), this LP, along with all associated click information, is fed into the same segmentation network for high-resolution refinement. The resulting refined LP is then pasted back into the corresponding position of the coarse mask $M_{\textnormal{c}}$ to produce a refined mask $M_{\textnormal{r}}$. Finally, in step (e), a variance analysis is applied to the previous prediction $M_{\textnormal{p}}$ and the current refined mask $M_{\textnormal{r}}$, yielding the final output for this iteration $M_{\textnormal{cur}}$. The entire process then cycles back to step (a) for any subsequent clicks, using $M_{\textnormal{cur}}$ as the new baseline prediction.

\subsection{Clore: Interactive Pathology Image Segmentation}
\subsubsection{Click Simulation and Encoding}To better approximate annotator behavior and provide interaction-aware supervision, we simulate user clicks and transform them into features compatible with the backbone. In training, 1–24 positive points are sampled from target regions and 0–24 negative points from background areas following an exponentially decaying distribution. Additional clicks, up to three, are placed in misclassified regions using the strategy of Sofiiuk et al. \cite{sofiiuk2022reviving}, with a small probability of resetting clicks to mimic user removal. Clicks are encoded as small-radius disks, passed through a convolutional module, and summed with image features before being fed into the backbone.

\subsubsection{Local Patch Selection}To balance efficiency and accuracy, we propose an adaptive strategy for selecting informative LP. During training, LP is determined by target size: the whole-object bounding box is used for small objects, while boundary-near regions are sampled with certain probability for larger ones. The rectangle is further augmented with random scaling and translation to include both the target and surrounding context. During inference, LP is obtained using Algorithm~\ref{alg1}.

\begin{algorithm}
\caption{Local Patch Selection}\label{alg1}
\label{alg:v1_compact}
\KwIn{
    $\mathcal{M}_{c}$: Current coarse prediction mask \\
    $\mathcal{M}_{p}$: Previous mask \\
    $\mathcal{R}_{g}$: Global region constraints $(y_{min}, y_{max}, x_{min}, x_{max})$ \\
    $(y,x)$: Last click coordinates \\
    $\gamma$: Expansion ratio
}
\KwOut{$\mathcal{B}=(y_1,y_2,x_1,x_2)$: Refinement region bounding box}
\BlankLine
$\mathcal{D} \gets \mathcal{M}_{c} ~xor~ \mathcal{M}_{p}$\;
\uIf{$\mathcal{M}_{p} = \emptyset$}{
    $\mathcal{B} \gets \text{BoundingBox}(\mathcal{M}_{c})$\;
}
\Else{
    $\mathcal{C} \gets \text{LargestConnectivityDomain}(\mathcal{D}, (y,x))$\;
    $\mathcal{B}_d \gets \text{BoundingBox}(\mathcal{C})$, $\mathcal{B}_o \gets \text{BoundingBox}(\mathcal{M}_{c})$\;
    \uIf{$\frac{\|\mathcal{B}_d\|}{\|\mathcal{B}_o\|} < \frac{1}{3}$}{
        $l \gets \max(\text{height of } \mathcal{B}_o, \text{width of } \mathcal{B}_o)$\;
        $\mathcal{B} \gets \text{Square with center at } (y, x) \text{ and side length } 0.4l$;
    }
    \Else{$\mathcal{B} \gets \mathcal{B}_d$\;}
}
$\mathcal{B} \gets \text{Expand }\mathcal{B}\text{ with the ratio } \gamma$\;
$\mathcal{B} \gets \mathcal{B}\cap \mathcal{R}_{g}$\;
\Return $\mathcal{B}$
\end{algorithm}

\subsubsection{Local Refinement} 

The Local Refinement stage is activated only after the predefined $n$ clicks, leveraging the specific prior information conveyed by later-stage interactions. The selected Local Patch (LP), along with all clicks within it, is fed into the shared segmentation network to generate the refined mask. A variance analysis is then performed between the refined mask and the previous mask. Specifically, we identify the areas of disagreement between the refined mask and previous masks, and select the connected component that contains the latest click. This component is then either added to or removed from the mask. This selective update mechanism ensures targeted correction of critical regions while preserving globally consistent predictions.

\subsubsection{Shared Segmentation Network} In contrast to methods using independent sub-networks for refinement segmentation, our approach employs a shared network for both Coarse Segmentation and Local Refinement. The shared parameters allow the refinement phase to leverage high-level semantic features from the coarse segmentation phase. And the shared architecture enables end-to-end joint optimization, improving feature consistency and allowing adaptive fusion of global and local information. Additionally, parameter reuse reduces model size, mitigating overfitting while maintaining accuracy. We use the HRNet+OCR~\cite{wang2020deep,yuan2020object} as the backbone, which has demonstrated strong performance in interactive segmentation tasks.

\subsection{Model Training}
Our training process differs slightly from the inference pipeline to enhance both computational efficiency and model robustness. We employ a hybrid strategy that combines random and iterative click simulation to prepare the input state of the model.
Specifically, before training, a gradient-free simulation stage is initiated to generate diverse and realistic segmentation states. We randomly sample 1-24 positive and 0-24 negative clicks on the input image and perform 0-3 corrective iterations. This strategy significantly improves model robustness, and the resulting mask is used as the `Previous Mask' input for subsequent, gradient-based training. The core training phase involves a single forward-backward pass, taking the `Previous Mask' and a single simulated corrective click as input. It performs a complete coarse-to-local refinement process, enabling us to supervise both coarse global prediction and high-fidelity local refinement simultaneously. This avoids the costly step of cropping separate image patches for extra training, thereby reducing computational overhead and enabling the shared network to concurrently learn both coarse and fine-grained abilities.

\subsection{Loss Functions}
The loss function consists of two parts: global loss for the coarse global segmentation, and local loss for Local Refinement. We provide auxiliary outputs in both stages to guide the model in learning richer semantic information. The global loss and local loss both use Normalized Focal Loss($\mathcal{L_\text{NF}}$)~\cite{lin2017focal,sofiiuk2019adaptis}, with Binary Cross-Entropy Loss($\mathcal{L_\text{BCE}}$) as the auxiliary. The formulas for both global loss and local loss are as follows:
\begin{equation}
\mathcal{L} = k_1 \cdot \mathcal{L_\text{NF}}+k_2 \cdot \mathcal{L_\text{BCE}},
\end{equation}
$k_{1}$ and $k_{2}$ are manually set weight coefficients, in our training, $k_{1}$ is set to 1 and $k_{2}$ is set to 0.4. The formula of $\mathcal{L_\text{NF}}$ is as follows:

\begin{equation}
\mathcal{L_\text{NF}}(g,p) = -\alpha \cdot \beta' \cdot \log(\min(p' + \epsilon, 1)).
\end{equation}

where \(g\) is the ground truth, \(p\) is the prediction, and \(p' = 1 - |g - p|\) measures the prediction accuracy. The term \(\beta' = \beta \cdot m\) is the normalized modulating factor, with \(\beta = (1 - p')^\gamma\) focusing on hard samples and \({m}\) balancing modulation strength. The weight \(\alpha\) can effectively address class imbalance, and \(\epsilon\) ensures sufficient numerical stability.

\begin{table*}[!t]
\centering
\caption{Quantitative Comparison of Segmentation Performance on Four Datasets}
\label{tab:performance}
\footnotesize
\renewcommand{\arraystretch}{1.3}
\setlength{\tabcolsep}{5pt}

\begin{tabular}{@{}lcccccccccccc@{}}
\toprule
\multirow{2}{*}{Dataset} & 
\multicolumn{4}{c}{RITM} & 
\multicolumn{4}{c}{FocalClick} & 
\multicolumn{4}{c}{Clore (Ours)} \\
\cmidrule(lr){2-5} \cmidrule(lr){6-9} \cmidrule(l){10-13}
& NoC@85 & NoC@90 & NoF@85 & NoF@90 
& NoC@85 & NoC@90 & NoF@85 & NoF@90 
& NoC@85 & NoC@90 & NoF@85 & NoF@90 \\
\midrule
Glas       & 3.18 & 6.55 & \textbf{1} & 7 & 3.33 & 5.21 & 3 & 4 & \textbf{2.48} & \textbf{4.15} & \textbf{1} & \textbf{3} \\
Nucls      & 18.67 & 19.83 & 27 & 29 & 15.90 & 18.30 & 22 & 26 & \textbf{12.90} & \textbf{16.50} & \textbf{17} & \textbf{22} \\
Digestpath & 18.70 & 19.58 & 45 & 47 & 18.18 & 19.74 & 39 & 47 & \textbf{16.46} & \textbf{18.94} & \textbf{32} & \textbf{45} \\
Bloodcell  & 1.43 & 5.43 & 5 & 62 & \textbf{1.31} & \textbf{2.21} & 2 & \textbf{8} & 1.33 & 2.68 & \textbf{1} & 10 \\
\bottomrule
\end{tabular}
\end{table*}

\begin{figure*}[t]
    \centering
    \includegraphics[width=0.7\linewidth]{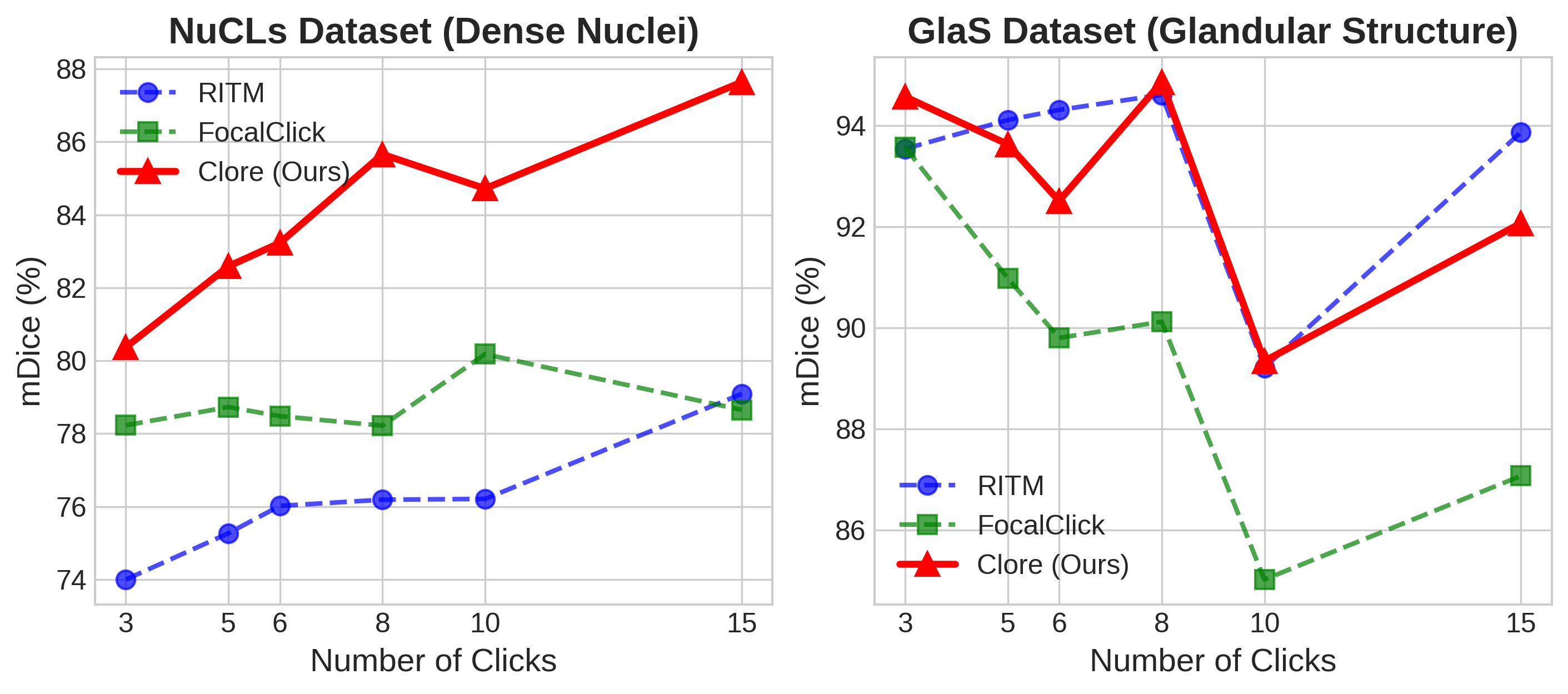}
    \caption{mDice performance on NuCLs and GlaS datasets at fixed number of clicks.}
    \label{fig:mdice}
\end{figure*}

\section{Experiments}
\label{sec:exp}

\subsection{Data Preparation}

We conduct experiments on four pathology image datasets: 1) DigestPath~\cite{da2022digestpath}: 250 H\&E-stained colorectal cancer images with pixel-level annotations of malignant lesions; 2) GlaS~\cite{amgad2019structured}: 151 H\&E-stained colorectal gland images annotated for both normal and cancerous glands; 3) NuCLS~\cite{kumar2019multi}: 44 multi-organ (breast/lung cancer) nucleus segmentation images containing densely distributed heterogeneous nuclei; and 4) BloodCell: 1,328 online-collected blood cell images with diverse staining protocols and exhaustive cell-boundary annotations. To address the computational challenges posed by high-resolution Whole Slide Images (WSIs), we strategically cropped annotated regions into smaller sub-images. This patch-based processing strategy not only facilitates efficient model training under memory constraints but also ensures that fine-grained local details are preserved for precise segmentation. We rigorously partition the entire dataset into three distinct subsets: training set (60\%), validation set (20\%), and test set (20\%).

\subsection{Implementation Details and Evaluation Metrics}

During training, the images and masks are first scaled up or down according to a scaling factor which is randomly generated from a truncated normal distribution. Here, we use a truncated normal distribution with a mean value of 0.8, a standard deviation of 0.4, a minimum value of 0.3, and a maximum value of 1.3. The images are then padded or cropped to a uniform size of 512×512 and undergo a random horizontal and vertical flip with a probability of 30\%. During validation, the images are only padded or cropped to the same uniform size of 512×512. For all experiments, we use the Adam optimizer with $\beta_1$ = 0.9 and $\beta_2$ = 0.999, and train the model for 230 epochs. The learning rate is initially set to $5 \times 10^{-4}$ and is reduced by a factor of 10 at the 20th and 40th epochs. Our network is implemented using PyTorch and trained and inferred on an NVIDIA H100 GPU with 80GB of memory. Based on the consideration of the generalization advantage of pre-trained weights, we used ImageNet-pretrained~\cite{deng2009imagenet} HRNetV2 as our shared segmentation network. Our method calculates the mean Number of Clicks (NoC) needed to achieve the specified Intersection over Union (IoU), along with the Number of samples that Fail to reach the target IoU (NoF) within the maximum number of clicks. The target IoUs are set to 85\% and 90\%, with a maximum of 20 clicks. The mean number of clicks exceeding 20 is therefore not precisely calculated. Lower values are generally preferred for both metrics in the evaluation.

\begin{figure*}[t]
    \centering
    \includegraphics[width=0.97\linewidth]{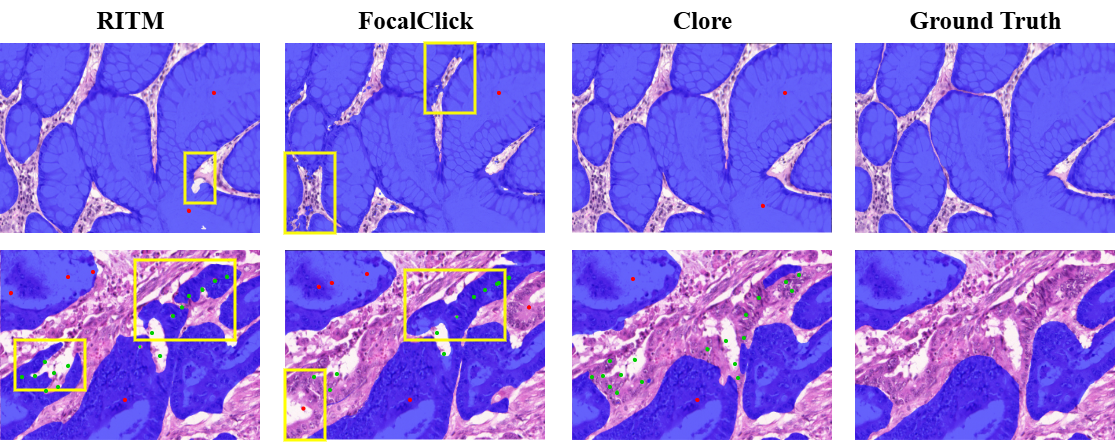}
    \caption{The visualized segmentation results of Glas dataset, where the upper row representing few click situation and the lower row representing many click situation. The yellow box indicates the better segmentation area of our method compared to the comparison method.}
    \label{figvis}
\end{figure*}


\subsection{Results}

Our method was evaluated against RITM and FocalClick on \textbf{four} datasets, with local refinement triggered at 5 clicks ($n = 5$). As shown in Table~\ref{tab:performance}, Clore achieves state-of-the-art NoC@90\% performance on GlaS, NuCLS, and DigestPath, reducing average clicks by 1.8 compared to baselines. Beyond the interaction cost metric (NoC), we also evaluated the segmentation quality using the mean Dice Similarity Coefficient (mDice)~\cite{taha2015metrics} under fixed interaction budgets to assess the convergence speed and stability of different methods. Figure~\ref{fig:mdice} illustrates the performance curves with respect to the number of clicks. On the NuCLS dataset, which presents a more challenging scenario with significantly higher NoC@85/90 metrics, Clore demonstrates a comprehensive advantage. It outperforms FocalClick by 2.14\% at 3 clicks and widens the gap to 8.99\% at 15 clicks, proving its exceptional robustness for complex, dense nuclear structures. Conversely, on the relatively simpler GlaS dataset with lower interaction thresholds, Clore not only achieves the highest accuracy in early stages (@3 clicks) but also maintains superior stability during mid-stage refinement (@8 clicks). This confirms the strong global segmentation capability of our shared network and its reliability in sustaining performance without oscillation even when receiving numerous corrective clicks. These results validate that our global-local cooperation mechanism rapidly captures target structures while effectively correcting subtle errors as interactions increase. For GlaS, global feature reuse models gland spatial context, while local refinement resolves adhesion boundaries, mitigating interaction redundancy. Figure~\ref{figvis} presents the visualized segmentation results of this dataset, where it is evident that our method outperforms others in both Few Click and Many Click scenarios. Specifically, Clore produces sharper boundaries around adjacent glands compared to the over-smoothed predictions of RITM, while effectively avoiding the fragmented masks often observed in FocalClick. On NuCLS, global prediction provides nuclear distribution priors, reducing repetitive clicks for dense nuclei. For DigestPath, Clore adaptively optimizes low-confidence regions under mucin interference, demonstrating that our dynamic feature fusion can effectively suppress complex texture noise that typically confuses purely global models. However, Clore underperforms FocalClick and RITM on BloodCell: BloodCell's inconsistent staining propagates noisy priors during global prediction, exacerbating errors during refinement. These results underscore that segmentation efficacy depends critically on target spatial distribution and annotation quality. Future work should address adaptive refinement for noisy scenarios.

\begin{figure}[t]
    \centering
    \includegraphics[width=0.97\linewidth]{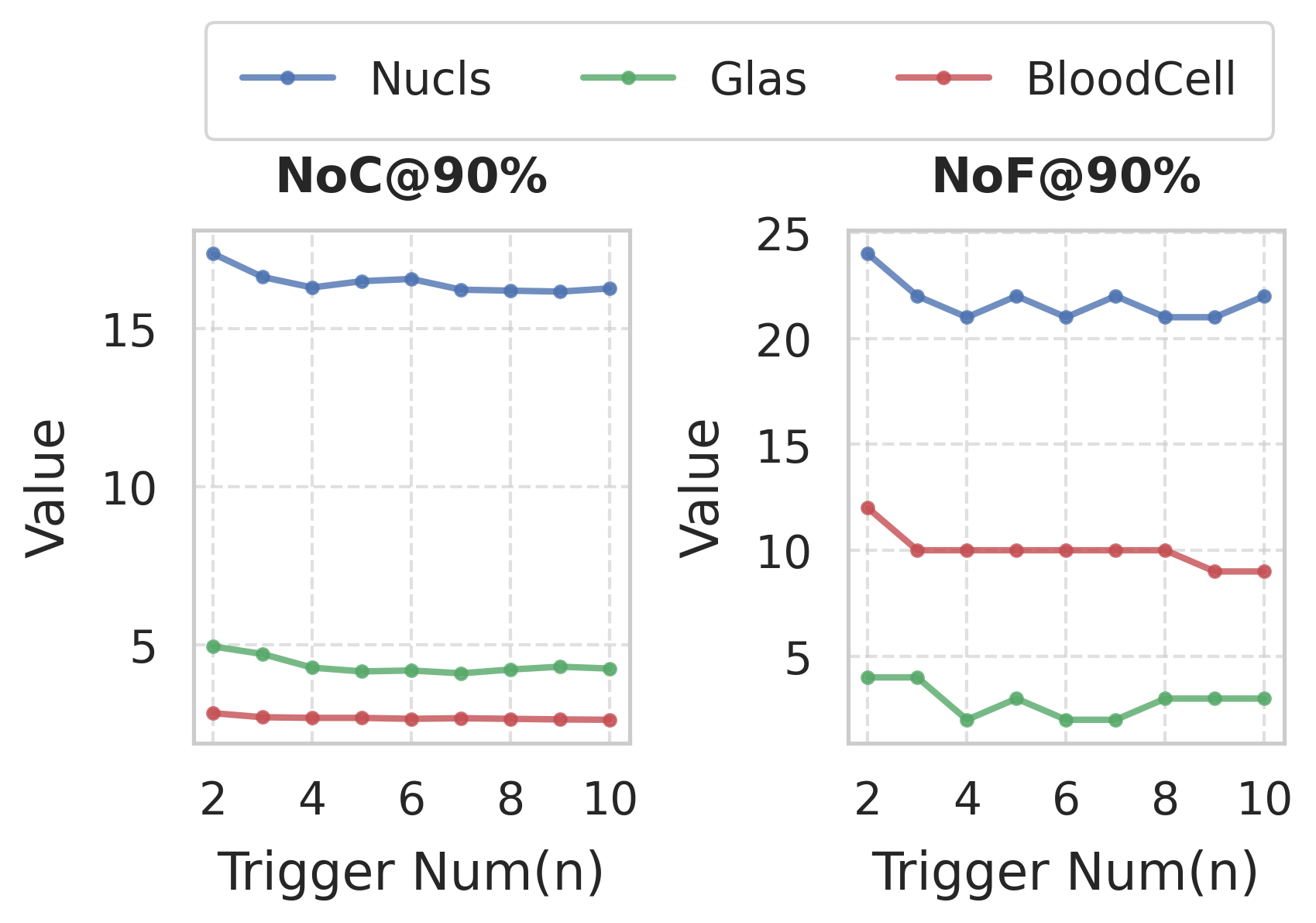}
    \caption{Result of the ablation study}
    \label{figablation}
\end{figure}

\subsection{Ablation Study}
To determine the optimal timing for triggering local refinement, we conducted an ablation study on the Nucls, GlaS, and BloodCell datasets. The experimental results, as illustrated in Figure \ref{figablation}, demonstrate that the number of clicks to trigger local refinement($n$) significantly impacts interaction efficiency. For the Nucls dataset (multi-organ dense nuclei), NoC@90\% stabilizes at 16.2 (±0.3) with $\geq5$ clicks, whereas premature triggering at 2-4 clicks fails due to insufficient global context. On the GlaS dataset, characterized by gland adhesion scenarios, setting the trigger at 5 clicks achieves the optimal balance, with a NoC@90\% value of 4.15, reducing the required clicks by 1.55 compared to triggering at 2 clicks. This improvement is attributed to the early localization of critical adhesion regions. Notably for BloodCell, which contains low-quality annotations, delaying refinement until $\geq7$ clicks improves performance by mitigating error accumulation in noise-sensitive contexts. The universal 5-click strategy balances global-local feature utilization across pathology images, demonstrating robust applicability despite dataset-specific characteristics.

\subsection{Efficiency Analysis}
We further evaluated the computational efficiency of different methods on the GlaS dataset, as summarized in Table~\ref{tab:efficiency_time}. We focused on two key metrics: Avg. SPC, representing the average seconds per click, and Total Processing Time, which measures the cumulative duration of model inference and necessary post-processing steps for annotating a sample. Existing methods often rely on heavy dual-network architectures or execute redundant coarse-to-fine re-predictions at every interaction, resulting in elevated latency. As shown in the table, the Avg. SPC of FocalClick reaches 125 ms. In contrast, Clore employs a shared segmentation network and triggers local refinement only when necessary. This strategy effectively reduces the per-click cost to 77 ms. Furthermore, benefiting from the significant reduction in the required number of interactions, our method achieves a Total Processing Time of 11 s, which is markedly faster than the 17 s and 23 s required by RITM and FocalClick, respectively. These results demonstrate that Clore offers a more efficient and practical solution for pathology annotation.

\begin{table}[!t]
\centering
\caption{Comparison of Inference Efficiency on Glas}
\label{tab:efficiency_time}
\footnotesize 
\renewcommand{\arraystretch}{1.3}
\setlength{\tabcolsep}{10pt}

\begin{tabular}{@{}lcc@{}}
\toprule
\multirow{2}{*}{Method} & 
\multicolumn{2}{c}{Computational Cost} \\
\cmidrule(lr){2-3}
& Avg. SPC (ms) $\downarrow$ & Total Time (s) $\downarrow$ \\
\midrule
RITM       & 87 & 17 \\
FocalClick & 125 & 23 \\
\textbf{Clore (Ours)} & \textbf{77} & \textbf{11} \\
\bottomrule
\end{tabular}
\end{table}

\section{Conclusion}
This paper introduces Clore, an interactive pathology image segmentation framework. Its core lies in a synergistic mechanism of "global prediction" and "local refinement", which dynamically balances segmentation efficiency and local precision via a hierarchical interaction strategy. The shared segmentation network and our training strategy also better combines multi-level image features while lightening the model. Experiments demonstrate Clore's superior performance in complex scenarios like gland adhesion and blurred boundaries. Although the framework faces limitations with severe staining noise and ultra-dense targets, its approach of aligning algorithmic design with domain characteristics offers valuable insights for advancing interactive pathology analysis. Future work will focus on enhancing the model's robustness and generalization in complex scenarios.

\bibliographystyle{IEEEtran}
\bibliography{mybib}
\end{document}